\documentclass{article}

\usepackage{microtype}
\usepackage{graphicx}
\usepackage{subfigure}
\usepackage[font=small]{caption}
\usepackage{times}
\usepackage{xspace}
\usepackage{times}
\usepackage{epsfig}
\usepackage{amsmath, bm}
\usepackage{amssymb}
\usepackage{algorithm}
\usepackage{algorithmic}
\usepackage{mathtools}
\usepackage{array}
\usepackage{multirow}
\usepackage[inline]{enumitem}
\usepackage{multirow}
\usepackage{tabularx}
\usepackage{color}
\usepackage[utf8]{inputenc} % allow utf-8 input
\usepackage[T1]{fontenc}    % use 8-bit T1 fonts
\usepackage[hyphens]{url}   % simple URL typesetting
\usepackage{booktabs}       % professional-quality tables
\usepackage{amsfonts}       % blackboard math symbols
\usepackage{xcolor}
\usepackage{blindtext}
\usepackage[numbers,sort]{natbib}
\usepackage{flushend}
\usepackage{booktabs}
\usepackage{lipsum}
\usepackage{bbm}

\usepackage{hyperref}

\usepackage[accepted]{icml2020}

\icmltitlerunning{Deep Adaptive Semantic Logic (DASL)}

\begin{document}

\twocolumn[ \icmltitle{Deep Adaptive Semantic Logic (DASL): Compiling
Declarative Knowledge into Deep Neural Networks}]

% It is OKAY to include author information, even for blind
% submissions: the style file will automatically remove it for you
% unless you've provided the [accepted] option to the icml2020
% package.

% List of affiliations: The first argument should be a (short)
% identifier you will use later to specify author affiliations
% Academic affiliations should list Department, University, City, Region, Country
% Industry affiliations should list Company, City, Region, Country

% You can specify symbols, otherwise they are numbered in order.
% Ideally, you should not use this facility. Affiliations will be numbered
% in order of appearance and this is the preferred way.
\icmlsetsymbol{equal}{*}

\begin{icmlauthorlist}
\icmlauthor{Karan Sikka}{sri}
\icmlauthor{Andrew Silberfarb}{sri}
\icmlauthor{John Byrnes}{sri}
\icmlauthor{Indranil Sur}{sri}
\icmlauthor{Ed Chow}{sri}
\icmlauthor{Ajay Divakaran}{sri}
\icmlauthor{Richard Rohwer}{sri}
\end{icmlauthorlist}

\icmlaffiliation{sri}{SRI International, USA}

\icmlcorrespondingauthor{Karan Sikka}{karan.sikka@sri.com}
\icmlcorrespondingauthor{John Byrnes}{john.byrnes@sri.com}

% You may provide any keywords that you
% find helpful for describing your paper; these are used to populate
% the "keywords" metadata in the PDF but will not be shown in the document
\icmlkeywords{deep learning, first order logic, knowledge, neural-symbolic, hybrid learning}

\vskip 0.3in

% this must go after the closing bracket ] following \twocolumn[ ...

% This command actually creates the footnote in the first column
% listing the affiliations and the copyright notice.
% The command takes one argument, which is text to display at the start of the footnote.
% The \icmlEqualContribution command is standard text for equal contribution.
% Remove it (just {}) if you do not need this facility.

\printAffiliationsAndNotice{}  % leave blank if no need to mention equal contribution
% \printAffiliationsAndNotice{\icmlEqualContribution} % otherwise use the standard text.

\def\@onedot{\ifx\@let@token.\else.\null\fi\xspace}
\makeatother
\def\etal{\emph{et al}\onedot}
\def\etc{\emph{etc}\onedot}
\def\ie{\emph{i.e}\onedot}
\def\eg{\emph{e.g}\onedot}
\def\cf{\emph{cf}\onedot}
\def\vs{\emph{vs}\onedot}
\def\pd{\partial}
\def\grad{\nabla}
\def\R{\mathbb{R}}
% I don't think this is used, and I overwrite it below:
% \def\L{\mathbb{L}}
\def\G{\mathbb{G}}
\def\d{\boldsymbol{\delta}}
\def\y{\textbf{y}}
\def\l{\boldsymbol{\ell}}
\def\wrt{w.r.t\onedot}
\def\a{\boldsymbol{\alpha}}
\def\vertspace{0.6em}
\newcommand{\mat}[1]{\bm{#1}}
\newcommand{\set}[1]{\mathbb{#1}}
\newcommand{\colons}[1]{``{#1}''}
\def\pytorch{PyTorch}

%Defining matrices and vector notations as per goodfellows book
\def\mF{\mat{F}}
\def\mX{\mat{X}}
\def\mS{\mat{S}}
\def\mM{\mat{M}}
\def\mW{\mat{W}}
\def\vs{\mat{s}}
\def\vx{\mat{x}}
\def\vt{\mat{t}}
\def\vh{\mat{h}}
\def\mA{\mat{A}}
\def\mD{\mat{D}}
\def\sE{\set{E}}
\def\sV{\set{V}}
\def\ssT{\mathbf{Triples}}
\def\sS{\mathbf{Unlabeled}}
\def\sY{\mathbb{Z}_{10}}
\def\sL{\mathbf{Labeled}}
\def\dvrd{\mathbf{D_\mathrm{vrd}}}
\def\imag{\mat{I}}
\def\conv{\circledast}

% Define logic and model theory symbols
\def\L{\mathfrak{L}}
\def\A{\mathfrak{A}}
\def\I{\mathcal{I}}
\def\D{\mathfrak{D}}
\renewcommand{\implies}{\rightarrow}
\def\loss{\mathcal{L}}
\newcommand{\logit}{\mathrm{logit}}
\def\digit{\mathrm{digit}}
\def\three{\mathit{three}}
\def\softselect{\mathit{softselect}}
\def\samp{\mathcal{S}}
\def\ie{\textit{i.e.}}
\def\eg{\textit{e.g.}}
\def\TRUE{\emph{True}}
\def\FALSE{\textit{False}}
\def\sT{\set{T}}
\def\sA{\set{A}}
\def\canride{\mathrm{CanRide}}
\def\riding{\mathrm{Riding}}
\def\vrd{\mathrm{vrd}}
\def\isridable{\mathrm{Ridable}}
\def\isliving{\mathrm{Living}}
\def\iswearable{\mathrm{Wearable}}
\def\wear{\mathrm{Wear}}

\definecolor{redcol}{rgb}{1, 0, 0}
\definecolor{bluecol}{rgb}{0, 0, 1}
\newcommand{\red}[1]{\textcolor{redcol}{#1}} 
\newcommand{\blue}[1]{\textcolor{bluecol}{#1}} 
\renewcommand{\paragraph}[1]{\smallskip\noindent{\bf{#1}}}

\def\algorithmautorefname{Algorithm}
\def\figureautorefname{Figure}
\def\tableautorefname{Table}
\def\equationautorefname{Eq.}
\def\sectionautorefname{Section}

\begin{abstract}
    We introduce {\bf Deep Adaptive Semantic Logic (DASL)}, a novel framework for automating the generation of deep neural
    networks that incorporates user-provided formal knowledge to improve learning from data.
    We provide formal semantics that demonstrate that our knowledge representation captures all of first order logic and
    that finite sampling from infinite domains converges to correct truth values.
    DASL's representation improves on prior neural-symbolic work by avoiding vanishing gradients, 
    allowing deeper logical structure, and enabling 
    richer interactions between the knowledge and learning components.
    We illustrate DASL through a toy problem in which we add structure to an image classification problem and demonstrate
    that knowledge of that structure reduces data requirements by a factor of $1000$. We then evaluate DASL on a visual
    relationship detection task and demonstrate that the addition of commonsense knowledge improves performance
    by $10.7\%$ in a data scarce setting. 
\end{abstract}

\section{Introduction}

Early work on Artificial Intelligence focused on Knowledge Representation and Reasoning (KRR) through the application of
techniques from mathematical logic \cite{genNilsAI}.  The compositionality of KRR techniques provides expressive power
for capturing expert knowledge in the form of rules or assertions ({\em declarative knowledge}), but they are brittle
and unable to generalize or scale.  Recent work has focused on Deep Learning (DL), in which the parameters of complex
functions are estimated from data \cite{lecun2015deep}.  DL techniques learn to recognize patterns not easily captured
by rules and generalize well from data, but they often require large amounts of data for learning and in most cases do
not reason at all \cite{yang2017differentiable, garcez2012neural, marcus2018deep, weiss2016survey}.  In this paper we
present \textbf{Deep Adaptive Semantic Logic (DASL)}, a framework that attempts to take advantage of the complementary
strengths of KRR and DL by fitting a model simultaneously to data and declarative knowledge.  DASL enables robust
abstract reasoning and application of domain knowledge to reduce data requirements and control model generalization.

\begin{figure}[tbp!]
  \begin{center}
	  \includegraphics[width=0.5\textwidth]{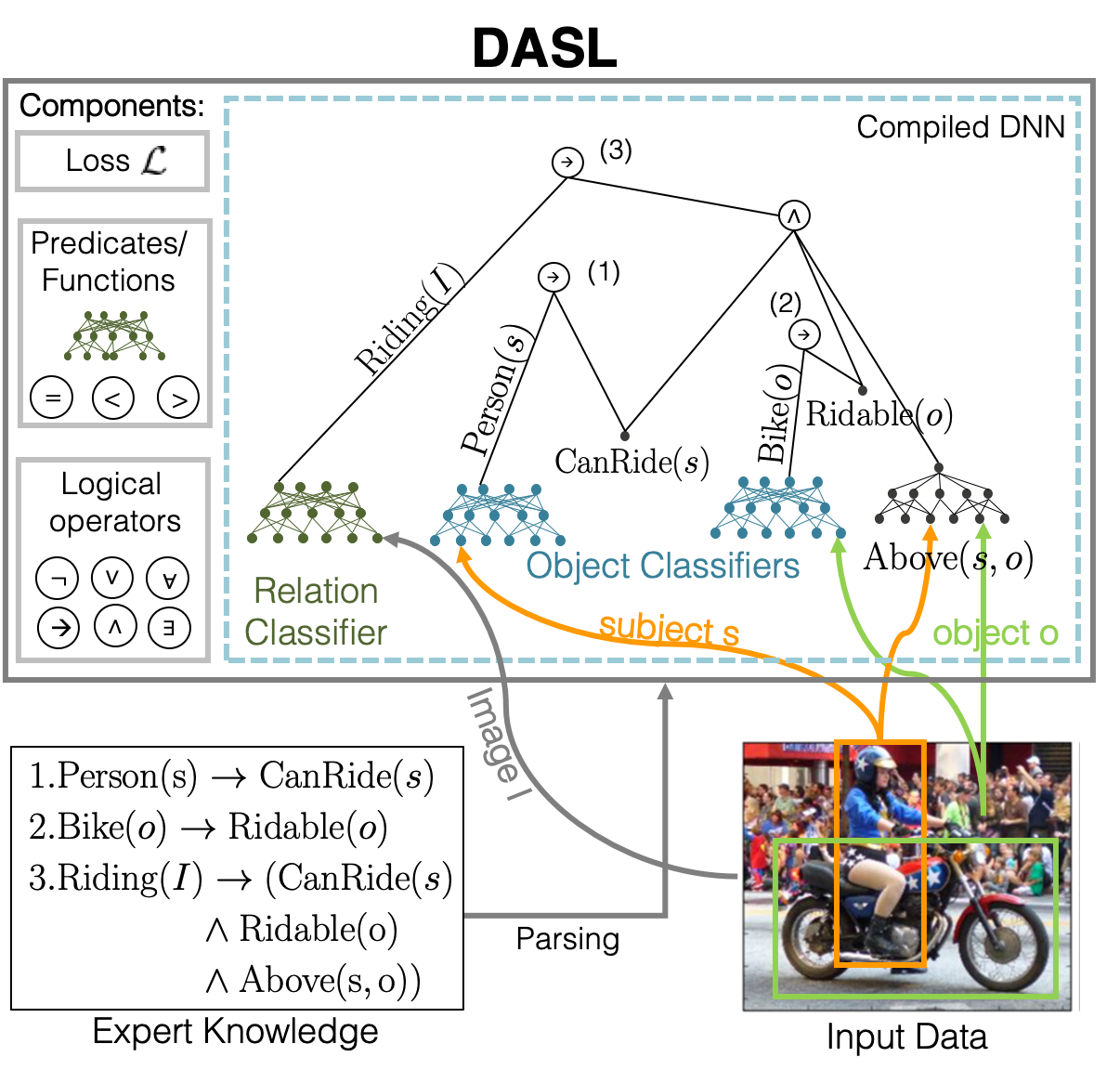}
  \end{center}
  \caption{
  DASL integrates user-provided expert knowledge with training data to 
  learn DNNs. It achieves this by compiling a DNN from knowledge, expressed in first 
  order logic, and domain-specific neural components. This DNN is trained using backpropagation, fitting
  both the data and knowledge. Here DASL applies commonsense knowledge  
  to the visual relationship detection task. $\wedge$ and $\rightarrow$
  refer to `and' and `implies' connectives respectively.}
  \label{fig:dasl}
\end{figure}

DASL represents declarative knowledge as assertions in first order logic.  The relations and functions that make up the
vocabulary of the domain are implemented by neural networks that can have arbitrary structure.  The logical connectives
in the assertions compose these networks into a single deep network which is trained to maximize their truth.
\autoref{fig:dasl} provides an example network that implements a simple rule set through composition of network
components performing image classification.  Logical quantifiers ``for all'' and ``there exists'' generate subsamples of
the data on which the network is trained.  DASL treats labels like assertions about data, removing any distinction
between knowledge and data.  This provides a mechanism by which supervised, semi-supervised, unsupervised, and distantly
supervised learning can take place simultaneously in a single network under a single training regime.

The field of neural-symbolic computing \cite{DBLP:journals/corr/abs-1905-06088} focuses on combining logical and neural
network techniques in general, and the approach of \cite{DBLP:journals/corr/SerafiniG16} may be the closest of any prior
work to DASL.  To generate differentiable functions to support backpropagation, these approaches replace pure Boolean
values of $0$ and $1$ for \TRUE{} and \FALSE{} with continuous values from $[0,1]$ and select fuzzy logic operators for
implementing the Boolean connectives.  These operators generally employ maximum or minimum functions, removing all
gradient information at the limits, or else they use a product, which drives derivatives toward $0$ so that there is
very little gradient for learning.  DASL circumvents these issues by using a logit representation of truth values, for
which the range is all real numbers.  %Products of truth values provide significant gradients in logit space.

Approaches to knowledge representation, both in classical AI and in neural-symbolic computing, often restrict the
language to fragments of first order logic (FOL) in order to reduce computational complexity.  We demonstrate
that DASL captures full FOL with arbitrary nested quantifiers, function symbols, and equality by providing a single
formal semantics that unifies DASL models with classical Tarski-style model theory \cite{chang1990model}.  We
show that DASL is sound and complete for full FOL.  FOL requires infinite models in general, but we show that
iterated finite sampling converges to correct truth values in the limit.  

In this paper we show an application of DASL to learning from small amounts of data for two computer vision problems.
The first problem is an illustrative toy problem based on the MNIST handwritten digit classification problem.  The
second is a well-known challenge problem of detecting visual relationships in images.  In both cases, we demonstrate that the
addition of declarative knowledge improves the performance of a vanilla DL model.  This paper makes the
following contributions:
\begin{enumerate}
  \item The novel framework DASL, which compiles a network from declarative knowledge and bespoke domain-specific reusable component
    networks, enabling gradient-based learning of model components;
  \item Grounding of the proposed framework in model theory, formally proving its soundness and completeness for full
    first order logic;
  \item A logit representation of truth values that avoids vanishing gradients and allows deep logical structures for neural-symbolic systems;
  \item Syntactic extensions that allow (i) restricted quantification over predicates and functions without violating first
    order logic constraints, and (ii) novel hybrid network architectures;
  \item Evaluation on two computer vision problems with limited training data, demonstrating that knowledge
    reduces data requirements for learning deep models.
\end{enumerate}

\section{Related Work}

\paragraph{Neural-Symbolic Computing:}
Early efforts to augment DNNs with logic focused on propositional logic, which supports only
logical connectives between (atomic) propositions
\cite{garcez2012neural, DBLP:journals/corr/abs-1905-06088}. For
example, KBANN \cite{Towell:1994:KAN:194414.194434} maps a set of
propositions into a graph, constructs a neural network,
and then trains it.
DASL follows this basic idea but fully supports full first order
logic (FOL) as well as arithmetic expressions.

Similar to several prior efforts 
\cite{hu2016harnessing, 
rocktaschel2015injecting, li2019augmenting}, DASL
replaces Booleans with real-valued \textit{pseudo-probabilities} 
to make the logical operations differentiable.
%so
%that the truth tables of the logical connectives can be replaced by
%differentiable interpolants of themselves.  
This circumstance has motivated the invention of 
a collection of {\em ad hoc} aggregation operators for representing
logical connectives \cite{Detyniecki_2001_AggOps}.  These include
the \textit{t-norm}, used by Logic Tensor Networks (LTNs)
\cite{DBLP:journals/corr/SerafiniG16} and the above works.  Instead, DASL uses a
logit representation for truth values, whose range is all real numbers, which avoids vanishing 
gradients and enables learning with deeper logical structures. 
DASL also differs in supporting multiple entity
types, arithmetic, and non-traditional operations such as {\em softmax}
that enable richer interaction 
between the NN and knowledge (\autoref{sec:experiments}).
DASL also represents the first time that 
soundness and completeness have been established 
for a FOL system applied to neural networks.

\paragraph{Compositional DL:} 
DASL is related to works that execute a task by composing trainable neural modules 
by parsing a query (in natural language)
\cite{AndreasRDK16, Yi0G0KT18, MaoGKTW19, yi2018neural}. 
For example, \cite{yi2018neural} focuses on visual question answering  
and employs a differentiable tree-structured logic representation, similar to DASL, but 
only in order to learn to translate questions into retrieval operations, 
whereas DASL learns the semantics of the application domain and can also integrate 
useful domain knowledge.

\paragraph{Structured Learning:} Other work also exploits underlying structure 
in the data or the label space to learn DNNs using techniques such as conditional 
random fields, graph neural networks, attention models, etc. \cite{belanger2017end, 
kim2017structured, battaglia2018relational, peng2018backpropagating, zheng2015conditional}. 
These methods impose structure
by either adapting the DNN architecture \cite{battaglia2018relational} or the loss function \cite{zheng2015conditional}.  
DASL instead imposes soft constraints by compiling DNNs based on rules that can  
be stated in a flexible manner using FOL. 

\paragraph{Weakly supervised learning:} DASL is related to works that use 
structural constraints as side-information or implicit knowledge to improve 
learning, particularly in data scarce conditions
\cite{xing2003distance, oquab2014weakly, 
chang2012structured,hu2016harnessing, NIPS2017_6969, stewart2017label}.

\paragraph{Semantic Reasoning:}
By the {\em semantics} of a logical language we mean an interpretation of its symbols (which do not include
logical connectives and quantifiers); a {\em model} in
the sense of model theory~\cite{Weiss97fundamentalsof}.  In common with several
methods~\cite{journals/corr/abs-1909-01161}, DASL grounds its entities in
vector spaces (embeddings) and its predicates and functions in trainable modules.
DASL builds on prior works on semantic representation
techniques \cite{pennington2014glove,NIPS2013_5021,dumaisLSI}
by enabling logical statements to modify the
entity embeddings so as to mirror semantic similarity in the application.

Traditional theorem provers \cite{siekmann} operate at a purely syntactic level to derive statements that hold true
regardless of the underlying semantics.  This approach often fails catastrophically when its users fail to supply
complete, accurate and consistent logical descriptions of their applications.  Approaches such as DASL that incorporate
semantic representations address this problem by treating the logic, like data, as merely suggestive.  An intermediate
approach~\cite{NIPS2017_6969, journals/corr/CohenYM17,journals/corr/abs-1906-06805} applies a theorem prover to a query
in order to generate a proof tree, which is then used to build a corresponding DNN.  Such methods can benefit from `soft
unification' in which proof steps can be connected via entities that nearly match semantically, rather than symbols that
match exactly or not at all.

\paragraph{Bayesian Belief Networks:}
Substitution of pseudo-probabilities for Booleans fails to capture uncertainty the way fully Bayesian methods do
\cite{jaynes03}. Bayesian Belief networks \cite{pearl2009probabilistic} accurately represent probabilities but lack
expressivity and face computability challenges.  Bayes nets are most comfortably confined to propositional logic.
Efforts to extend them to first-order logic include Markov Logic Networks~\cite{richardson2006markov}, which use an
undirected network to represent a distribution over a set of models, \ie, {\em groundings} or {\em worlds} that can
interpret a theory. The lifted inference approach \cite{Kimmig04liftedgraphical} reasons over populations of entities to
render the grounded theory computationally tractable.  These methods generally do not support the concept of
(continuous) soft semantics through the use of semantic embedding spaces, as DASL does.

\section{Approach}
\label{sec:approach}

In this section we describe our approach to integrate data with relevant expert knowledge.  Consider the task, depicted
in \autoref{fig:dasl}, of predicting the relationship between bounding boxes containing a subject and an object.  In
addition to learning from labeled training samples, we want to incorporate the commonsense knowledge that if the
predicted relationship is \colons{Riding} then the subject must be able to ride, the object must be ridable,
and the subject must be above the object.  Incorporating such knowledge results in a more robust model that uses
high-level semantics to improve generalization and learn from a small number of examples.  DASL achieves integration of
the continuous representations in DNNs with the discrete representations typically used for knowledge representation by
compiling a DNN from the knowledge assertions and grounding the vocabulary of the domain in component networks, enabling
gradient-based learning of the model parameters.

We begin by providing the theoretic underpinning of DASL in FOL.  We then describe the underlying representations of the
DASL model including: model components, language elements, etc., which ground the formal language and allow end-to-end
learning of model parameters.

\subsection{DASL Model Theory}

A DASL theory is specified in a language $\L$ containing constants $a_0,\ldots$, function symbols $f_0,\ldots$, and
relation symbols $R_0,\ldots$.  In addition, we have variables $x_0,\ldots$ understood to range over objects of some
universe, logical connectives $\neg$ (`not') and $\wedge$ (`and'), the quantifier $\forall$ signifying `for all', and
the single logical binary relation symbol `$=$' indicating equality. For presentation purposes we treat $\vee$ (`or'),
$\implies$ (`implies'), and $\exists$ (`there exists') as defined in terms of $\neg$, $\wedge$, and $\forall$ (although
they are implemented as first class connectives in DASL). Constants and variables are {\em terms}; an $n$-ary function
symbol applied to $n$-many terms is a {\em term}.  An $n$-ary relation symbol (including equality) applied to $n$-many
terms is a {\em formula}; if $\phi$ and $\psi$ are formulas and $x$ is a variable then $(\forall x)\phi$, $\neg\phi$,
and $\phi \wedge \psi$ are {\em formulas}.

Formal semantics for $\L$ are provided by structures interpreting the symbols of $\L$. We generalize the typical
Tarski-style \cite{Weiss97fundamentalsof} model theoretic definitions to capture DASL models.  A model
maps every term to an element of the domain and every formula to a {\em truth value}.  In classical semantics a model
maps every formula either to \TRUE{} or to \FALSE{}, frequently represented as $1$ and $0$.  To apply general
optimization techniques such as stochastic gradient descent (SGD), we define the truth values to be the closed real
interval $[0,1]$, denoted as $\sT$.

We begin by specifying a class of objects $A$, the {\em domain} of the model. A {\em variable map} $V$ for $A$ maps
variables in $\L$ into $A$. An {\em interpretation} for $\L$ and $A$ is a structure $\I = (I, \I_{f_0},\dots,
\I_{R_0},\ldots)$ such that $I$ maps constants into $A$, $\I_{f_i} : A^{m_i} \rightarrow A$ and $\I_{R_j}
: A^{n_j}
\rightarrow \sT$ for each $i$ and $j$ where $m_i$ and $n_j$ are the arities of functions $f_i$ and $R_j$. Given $\I$
and $V$ for $A$, $\A = (A, \I, V)$ is called a {\em model} for $\L$.

We use connectives to define functions on truth values.  For truth values $t_1, t_2, \ldots$ we define $\neg t_1 =
1-t_1$, $t_1 \wedge t_2 = t_1 \cdot t_2$, and $\forall_i t_i = \Pi_i t_i$.  We also allow different
definitions of these functions.  We interpret $=$ using a function $\D_=$ on objects $u,v \in A$ such that
$\D_=[u,v] \in \sT$ and is $1$ if and only if $u=v$.  Finally, we define a {\em sampling function} $\samp$ that maps the
domain $A$ to an arbitrary subset of $A$.

Given $\A = (A, \I, V)$, variable $x$, and $u \in A$,
$\A_{u/x}$ is the model $(A, \I, V^*)$ where $V^*(x) = u$ and $V^*(y) =
V(y)$ for $y$ other than $x$.  We now define
interpretation in $\A$ of variable $x$, constant $a$, term
$t_1,\ldots,t_n$, function symbol $f$, relation symbol $R$, and
formulas $\phi$ and $\psi$, all from $\L$, by the following inductive
definition:
\begin{eqnarray*}
\A[x] & = & V(x) \\
\A[a] & = & I(a) \\
\A[f(t_1,\ldots,t_n)] & = & \I_f(\A[t_1],\ldots,\A[t_n]) \\
\A[R(t_1,\ldots,t_n)] & = & \I_R(\A[t_1],\ldots,\A[t_n]) \\
\A[t_1=t_2] & = & \D_=(\A[t_1], \A[t_2]) \\
\A[\neg\phi] & = & \neg\A[\phi] \\
\A[\phi \wedge \psi] & = & \A[\phi] \wedge \A[\psi] \\
\A[(\forall x) \phi] & = & \forall_{u \in \samp(A)} \A_{u/x}[\phi]
\end{eqnarray*}

When $\A[\phi]=1$ we write $\A \models \phi$ and we say $\A$ is a model of $\phi$ and satisfies $\phi$.  If $\Gamma$ is
a set of formulas and $\A \models \phi$ for every $\phi \in \Gamma$ then we write $\A \models \Gamma$.

The standard semantics from model theory are achieved when the range of $\D_=$ and $\I_R$ is $\{0,1\}$ and when
$\samp(A) = A$. For basic DASL semantics, $A = \mathbb{R}^N$ for some fixed $N$.  DASL also extends to many-sorted logic;
\ie, when bound variables have types, the single universe $A$ is replaced by a universe $A_i = \mathbb{R}^{N_i}$ for
each sort and supported by the above definitions. We allow the sampling function $\samp(A)$ to return different samples on
different invocations. The mapping $I$ from constants to $A$ is referred to as \textit{embedding} (as done in deep
learning). A function $\loss$ that maps sequences of truth values to non-negative reals is a {\em DASL loss function} if
$\loss(\langle 1 \rangle)=0$ and $\loss$ is monotonic in every element of its input sequence.  We 
define $\A \models_\theta \Gamma$ whenever $\loss(\A[\Gamma]) \leq \theta$. Thus $\A \models_0 \Gamma$ is equivalent to  $\A \models
\Gamma$.

\paragraph{DASL is sound and complete:} To prove formally that DASL
models capture full first order logic, we show that for any set of
formulas $\Gamma$ there is a Tarski model $\A \models \Gamma$ if and
only if there is a DASL model $\mathfrak{B} \models \Gamma$.  By the definitions
provided, a DASL model can be constructed equivalent to any
given Tarski model by replacing objects by vectors and sets by
functions.  Since we do not restrict the class of functions, this
is trivial.  When a DASL model has $0$ loss for $\Gamma$, construction
of a Tarski model is straightforward as well.

The more interesting questions come when we restrict DASL to what is
computationally feasible and when we generalize to $\A \models_\theta
\Gamma$.  Suppose the domain $A$ of $\A$ can be expressed as $A_1 \cup
A_2 \ldots$ where the disjoint $A_i$ are all finite and of fixed
cardinality.  If $\loss(A_i \cup A_j) = \frac{1}{2}\loss(A_i) +
\frac{1}{2}\loss(A_j)$ for all $i \neq j$ then $\sum_i
\loss(\A[\Gamma, A_i]) = \loss(\A[\Gamma])$ (where we oversimplify
slightly by omitting the details of defining average loss over
infinite domains). Thus, even for a finite sampling function from an
infinitary domain, we compute correct loss in the limit when repeated
applications of the sampler yield $A_1, A_2, \ldots$.  When the
interpretation functions of $\I$ are implemented as neural networks, there
will be restrictions on the classes of functions and relations that
can be computed, and these have been well-studied. 

\subsection{DASL Models as Neural Networks}
\label{sec:approach-nn}

Given a DASL theory $\Gamma$, DASL searches for a model $\A$ that satisfies $\Gamma$ with minimal loss.  $\Gamma$ in
general contains both data for a standard machine learning task and knowledge assertions such as those in
\autoref{fig:dasl}.  We implement DASL in the popular deep learning library \pytorch{} \cite{paszke2019pytorch}
%, to leverage its functionality\footnote{Although we use \pytorch{} for implementing DASL, it can be implemented any other
%  deep learning library.} and speed to find $\A$
.  The DASL semantics defined above are both compositional and a
function of the syntax of $\Gamma$, at least down to choice of $\I$.  Since neural networks are also compositional, DASL
constructs independent networks for each function in $\I$ and assembles these into a single network based on
the parse tree of $\Gamma$. This makes DASL compositional, where DNNs are assembled on
the fly based on the theory. We then use backpropagation through the tree to minimize the loss to learn the model
parameters. We next describe details for the internal representations, implementation of language elements,
optimization, and extensions of logical language.

\paragraph{Representation of model components:} Implementation of a DASL model requires specification of a domain $A_i$ for each
logical type. The domains can include both parameterized members and static members, which can be continuous (\eg{} visual
embeddings), ordinal and categorical (\eg{} labels) types. For each domain $A_i$ having elements represented by constants,
we need to specify the embedding $I_i$ for the constants. Any neural network implementable in \pytorch{} can be
used to realize $\I_f$ and $\I_R$.

DASL works with logits rather than truth values.
The logit for a truth value $t \in \sT$ is calculated as $\logit(t)= \ln\frac{t}{1-t}$ and its
inverse is a sigmoid non-linearity ($t = \sigma(\logit(t))$).

\paragraph{Implementation of the logical connectives:}
For truth values $t_1$ and $t_2$ and corresponding logits $l_1$ and $l_2$, we define negation ($\neg$) and conjunction
($\wedge$) operators as:
\[
%\neg l_1 = \logit(1-t_1)\\
\neg l_1 = \logit(1-t_1) = -l_1\\
\]
\[
l_1 \wedge l_2 = \logit(t_1t_2) = \ln\sigma(l_1) + \ln\sigma(l_2) - \ln(1 - \sigma(l_1)\sigma(l_2))
\]

This formula for $\wedge$ is numerically unstable when $t_1t_2$ gets close to $1$.
Whenever this occurs, we instead use the numerically stable approximation:
\[
l_1 \wedge^* l_2 \approx -\ln(e^{-l_1} + e^{-l_2}).
\]
We use \pytorch{} functions {\tt logsigmoid} and {\tt logsumexp} that provide efficient and numerically robust
computations for terms arising in these equations.

Conjunction and universal quantification are naturally represented as products of truth values, but the product of a
large number of positive terms all less than $1$ gets arbitrarily close to $0$, and so does its derivative, meaning that
learning is slow or will stop altogether.  Under the logit space equations above, however, conjunctions are sums, so
increasing the number of conjuncts does not diminish the gradient.  Two typical alternatives for $t_1 \wedge t_2$ in
systems that operate directly on truth values are $\min(t_1, t_2)$ and $\max(0, t_1+t_2-1)$
\cite{DBLP:journals/corr/SerafiniG16}. When many terms are conjoined, the former formula yields gradient information
only for the minimal truth value, and the second yields no gradient information at all whenever $t_1 + t_2 < 1$, again
restricting the ability of a system to learn.

\paragraph{Equality:}
DASL functions can include standard regression tasks $f(x)=y$ for
real-valued $y$.  The behavior of DASL on such rules is governed by
$\D_=(f(x),y)$, so $\D_=$ needs to be a function that allows for
backpropagation.  Since we reason in logit space, $\D_=$ cannot be
implemented as mean squared error since its logit would rapidly diverge
towards infinity as the error gets small.
Instead we take the logit transform
to be a log likelihood $\ln(\frac{Pr(u=v)}{Pr(u\neq v)})$ and we model $d=u-v$ as
normally distributed noise when $u$ and $v$ are ``equal'' (with mean 0
and variance $\varepsilon^2$) and as
normally distributed distance when $u$ and $v$ are genuinely
different (with mean $\mu$ and variance $\sigma^2$).
Ignoring the scaling factor, the density for
$x=|d|$ in the latter case is given by
$e^{-(x-\mu)^2/2\sigma^2} + e^{-(x+\mu)^2/2\sigma^2}$.
Using the ratio of these densities in place of the ratio of
probabilities, we derive:
\(
\logit(\D_=(u,v))=
\ln \frac{2\sigma}{\varepsilon} + \frac{x^2}{2\varepsilon^2}
- \ln (e^{-(x-\mu)^2/2\sigma^2}
       + e^{-(x+\mu)^2/2\sigma^2})
\)
When $u$ and $v$ are vectors rather than scalars, we can use the same
distribution on $||u-v||$.

\paragraph{Quantifiers and sampling:}
As mentioned previously, the sampler may return different samples on different invocations. A sampler is implemented as
a \pytorch{} {\tt dataloader}, so returned samples are similar to mini-batches in SGD.  The types of quantified variables are
always specified, and may be drawn from a fixed table of values (such as images), a fixed table of parameterized values
(vectors to be learned), or all vectors in a vector space. A sampler is defined and can be customized for each
quantifier instance, so that different quantifiers over the same type can sample differently.  When quantifiers are
nested, samples obtained by outer samplers are available as input to inner samplers. For example, in $(\forall
x:T_1)(\exists y:T_2)\phi$, the sampler which selects some set of $y$'s from $T_2$ may rely on $x$ to determine which
$y$'s to sample.  In this sense, the samplers are similar to Skolem functions \cite{hodges1997shorter}.  Because samples
are always finite, $\forall$ is implemented as the product of all elements of the sample.

\paragraph{Optimization:} DASL replaces $\Gamma$ with the conjunction of all of its elements and thus the loss function is
applied to a single truth value.  We define $\loss(t)$ as the cross-entropy between the distributions $(t, 1-t)$ and $(1,
0)$, which yields $\loss(t) = -\ln(t) = \ln(1+e^{-l})$, where $l$ is the logit of $t$.  Not only is this loss function
known to perform well in general on binary data, but together with our interpretations of the logical connectives
it satisfies the condition above for sampling to converge to the total loss (even under infinite domains).

\paragraph{Extending the logical language:}
We describe the implementation of equality above; less than and greater than are implemented similarly.  We do not
require functions to be implemented as learned neural networks; they can be coded deterministically in \pytorch{} if
desired.  Several arithmetic operations are incorporated into the language in this way. We further extend the DASL language
to allow for convenient specification of efficient networks.  Firstly, connectives automatically operate on arbitrary
sequences of logits.  For example, $\wedge(u_0,\ldots,u_n) = u_0 \wedge u_1 \wedge \ldots u_n$.  The connectives
also generalize to work component-wise on tensors and support broadcasting as is familiar for tensor operators in
\pytorch{}. For example, if $\mat{X}$ is a matrix and $\mat{y}$ is a vector, both of logits, then $\mat{X} \wedge
\mat{y} = \mat{Z}$, where $Z_{ij} = X_{ij} \wedge y_{i}$.

The above property makes it possible to conveniently express knowledge as vectors of formulas, which can take
advantage of tensor operations in \pytorch{}.  We use this ability in \autoref{sec:vrd} to reduce the learning
requirements on a classifier $\mathrm{classify}(x)$ that maps input $x$ to a value per class; these values would
typically then pass to a softmax operation.  We know that certain classes $A$ could only be correct under conditions
$\phi$, which are detected outside the classifier, so we write $\mathrm{classify}(\vx) \wedge (\mathrm{AClasses}
\rightarrow \phi(\vx))$ where $\mathrm{AClasses}$ is a constant vector over all classes with value $1$ for the classes
which are in $A$ and $0$ otherwise.  The effect of this operation is to mask the output of the classes in $A$ whenever
$\phi$ does not hold.  Since the ground truth label will be compared to the output of the $\wedge$ node, the classifier
only receives feedback on these classes when $\phi$ holds, which is the only time it could ever receive positive
feedback.  The overall system is capable of learning to generate correct labels, while the classifier itself does not
have the burden of learning to suppress the classes in $A$ when $\phi$ does not hold.  Boolean vectors are implemented
by defining $\logit(1)$ to be a fixed large constant.

Finally, we provide an explicit operator $\softselect(\Gamma, i)$ (denoted as $\pi_i(\Gamma)$) which outputs the logit
value for the $i^\mathrm{th}$ formula of $\Gamma$ after application of the logit version of the softmax operator. This
allows us to directly specify standard architectures for multi-class classification problems and to allow rules to
operate on the classifier output within the model. Because $i$ is an integer argument, we can quantify over it,
effectively quantifying over a fixed finite list of predicates, providing syntactic convenience without violating the
constraints of FOL.

\section{Experiments} 
\label{sec:experiments}

We evaluate DASL on two computer vision problems in data scarce conditions.
We show that DASL augments 
deep learning with declarative knowledge to achieve better generalization.
The first task is a toy problem based on digit classification on the MNIST 
dataset \cite{mnist}, where knowledge is provided as an arithmetic relation
satisfied by unlabeled triplets of digit images that are arranged artificially
to satisfy that relation (\autoref{sec:mnist}).  
We then focus on the problem of detecting visual
relationships between object pairs and use commonsense knowledge 
about the plausible arguments of the relationship (\autoref{sec:vrd}). 

\subsection{Toy Example on MNIST} 
\label{sec:mnist}

\paragraph{Problem statement:}  
We use this toy example to demonstrate DASL's ability to
train a NN from a few labeled samples and large number of unlabeled
samples satisfying a rule. 
We denote a grayscale input image of a MNIST digit as $\mX$
and its label (if provided) as $y (\mX) \in \sY$, where
$\sY = \{0, 1, ..., 9\}$. The task is to learn a NN $\digit(\mX)$ to
predict the digit in a test image.

For our toy example, we split the training data ($50K$ images) into
two disjoint sets: $\sL$, containing a small number $N_{tr}$ of
labeled examples per digit class, and $\sS$, used to generate the
set $\ssT$ containing triplets of images $(\mX_1,\mX_2,\mX_3)$
satisfying the rule $y(\mX_1)+y(\mX_2)=y(\mX_3)\bmod{10}$.
$\ssT$ contains only unlabeled images that together satisfy this relationship.
We wish to learn the classifier by using $\sL$ and $\ssT$,
and thus the challenge is to compensate for the small size of $\sL$
by leveraging the prior knowledge about how the
unlabeled images in $\ssT$ are related.
% Our experiments also shows how DASL can use the supervision available from prior knowledge to learn a neural network.
We formulate this problem within DASL by using its $\softselect$ operator $\pi_i$ (see \autoref{sec:approach-nn}) that, applied to the NN output
$\digit(\mX)$, returns the normalized score for the $i^{th}$ class.  This rule is written:
% \vspace{-2em}
\begin{align*}
  & (\forall (\mX_1,\mX_2,\mX_3):\ssT) (\forall y_1:\sY) (\forall y_2:\sY)   \\
	& \qquad [(\pi_{y_1}(\digit(\mX_1)) \wedge \pi_{y_2}(\digit(\mX_2))) \\
  &	\qquad\qquad\qquad\qquad
    \rightarrow \pi_{(y_1\hspace{0.5ex}+\hspace{0.5ex}y_2)\bmod 10}(\digit(\mX_3))]
\end{align*}

\begin{figure}[htbp!]
	\begin{center}
		\includegraphics[scale=0.36]{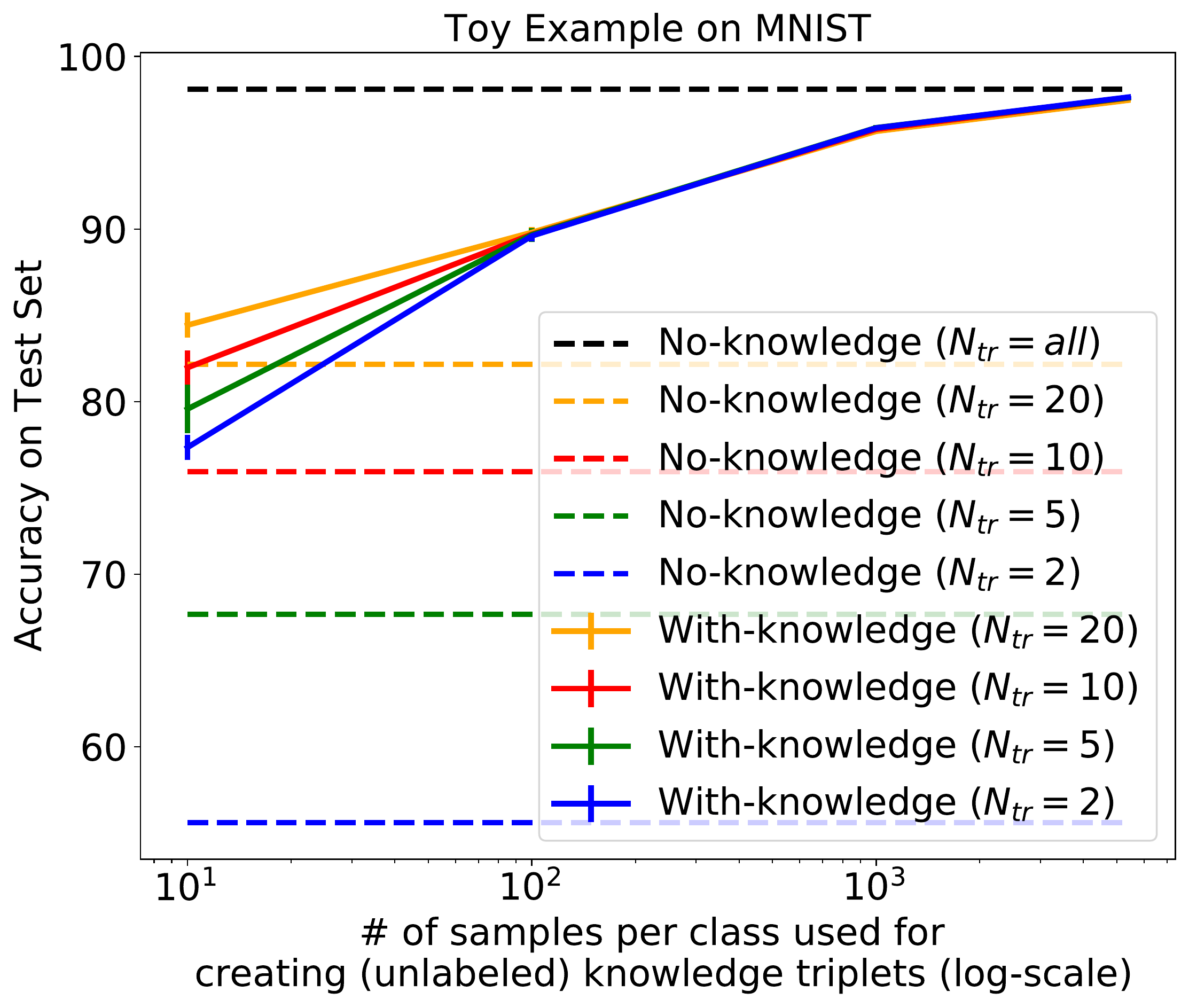}
	\end{center}
	\caption{Figure showing the results for the MNIST toy example with a plot of
    accuracy of digit classification versus number of 
    samples per class used for creating the unlabeled knowledge triplets.
    The labels \textit{With-knowledge} and \textit{No-knowledge} denote
    whether the training included the
    knowledge-augmented unlabeled triplets satisfying the
    given modular arithmetic (see \autoref{sec:mnist}).
    $N_{tr}$ refers to the number of labeled training
		examples per class (\textit{all} 
    refers to the entire training set).
    Best seen in color.
	}
	\label{plot:mnist}
  \end{figure}

We quantify over the triplets from $\ssT$ and all possible pairs of digits from $\sY$.  We use this theory to augment
the theory corresponding to the labeled training examples $(\forall(\mX):\sL) (\pi_{y(\mX)}(\digit(\mX)))$.
%as the $\implies$ will become
%\FALSE{} only when the antecedent is \TRUE{} and consequent is \FALSE{}.  This is not a trivial learning problem since
The model is required to correctly infer the (unknown) labels of the triplet members and then use them for indirect
supervision.  We evaluate the model using the average accuracy on the test set ($10K$ images).  For $\digit(\mX)$, we
used a two-layer perceptron with $512$ hidden units and a sigmoid non-linearity.  We performed experiments in data
scarce settings with $N_{tr}=2, 5, 10$, and $20$, and report mean performance with standard deviation across
$5$ random training subsets as shown in \autoref{plot:mnist}.  We use an equal number of examples per-class for
constructing the triplets.  We use a curriculum based training strategy (see supplementary) to prevent the model from
collapsing to a degenerate solution, especially for lower values of $N_{tr}$.  We train the model with the Adam
optimizer \cite{kingma2014adam}, learning rate of $5 \times 10^{-5}$, and batch size of $64$. We report performance
after $30K$ training iterations.  A test image is classified into the maximum scoring class.

\paragraph{Results:}
\autoref{plot:mnist} shows a plot of digit classification accuracy
versus the number of samples per class used for
creating the triplets. We observe that the NN trained with both knowledge
and data (\textit{With-knowledge}) outperforms its counterpart
trained with only labeled samples (\textit{No-knowledge}). 
The improvement is particularly notable when training with smaller
labeled training sets; \eg{}, for $N_{tr}=2$, using all the knowledge
raises performance from $53.3 \pm 1.01$ to $97.7 \pm 0.00$.
We also note that the performance of the \textit{With-knowledge}
model improves as the number of triplets increases and converges to 
similar values for different values of $N_{tr}$, indicating that the knowledge renders
extra labels largely superfluous.  The mean performance is $97.6 \pm 0.00$,
which is competitive with the performance of a model trained with all $50K$
labeled examples in MNIST ($98.1$ for $N_{tr}=all$).
These results demonstrate the strength of DASL for exploiting
knowledge to dramatically reduce data requirements.
It also shows how DASL optimizes NNs that represent the domain language,
using backpropagation to fit data and knowledge.

\subsection{Visual Relationship Detection} 
\label{sec:vrd}
Many problems in machine learning are endowed with inherent structure that can often be described explicitly.  We show
this in the visual relationship detection task, where DASL incorporates commonsense knowledge into a DNN to improve
learning with a small amount of training data.

\paragraph{Problem Statement:} We use the Visual Relationship Detection (VRD) benchmark \cite{lu2016visual} 
to evaluate the \textbf{Predicate Detection Task:} Given an image and a set of ground-truth bounding boxes with object
category labels, predict the predicates that describe the relationships between each pair of objects. The VRD dataset
contains $5000$ images spanning $37993$ relationships covering $100$ object classes and $70$ predicate classes. We use
splits provided by the authors that contain $4000$ train and $1000$ test images. The dataset also provides a
\textit{zero-shot} test subset of $1877$ relationships built from the same classes as the training data but containing
novel combinations of predicate classes with object class pairs.

\paragraph{Baseline model:} 
We begin with a NN $\vrd(\imag, s, o)$ that outputs raw scores for predicate classes, where $\imag$ is the input RGB
image and $s$ and $o$ are the indices of the subject and object classes respectively.
We implement two variants of $\vrd$ similar to that proposed in \cite{liang2018visual}.  The first variant, referred to as
\textit{VGG}, extracts visual features from the last layer of a pre-trained VGG-16 network from the bounding box of the
subject, the object, and their union.  These features are projected into a $256$ dimensional space by using a projection
layer $P$ (made of a fully-connected (FC) layer and a ReLU non-linearity) and then fused by concatenation. The features
are passed through another $P$ layer followed by a FC layer to predict the class-wise scores. The second variant,
referred to as \textit{VGG-SS}, additionally incorporates the word-embedding features of the subject and the object
($300$ dimensional Glove features \cite{pennington2014glove}) along with the normalized relative spatial coordinates
(see supplementary).  These features are first projected using additional $P$ layers and then concatenated with visual
features, as done for VGG, prior to predicting the class-scores.  We train the model with Adam optimizer
\cite{kingma2014adam}, learning rate of $5 \times 10^{-5}$, and batch size of $128$.

\paragraph{DASL based Approach:} We deviate from the simplified model of \autoref{fig:dasl}, instead expressing knowledge 
as vectors of formulas as discussed in \autoref{sec:approach-nn}.  We begin by defining $\canride$ as a constant vector
of truth values for all objects which is \TRUE{} at indices of objects which can ride and \FALSE{}
elsewhere. $\canride(s)$ selects its $s^\mathrm{th}$ element.  Similarly, we define $\isridable$ as a vector which is \TRUE{} at
exactly the indices of objects which can be ridden. Finally we define a one-hot vector of truth values $\vh_{Cls} \in \R^{70}$,
which is \TRUE{} at the index of the predicate class \colons{Cls} and \FALSE{} elsewhere.  The theory which asserts that
$\vrd$ should output the class labels assigned in the training data and that the \colons{Riding} predicate should only apply
when the subject can ride and the object can be ridden is written as:
\begin{align*}
	&(\forall (\imag, s, o, y):\dvrd) [\pi_y(\vrd(\imag, s, o) \\ 
	& \quad \wedge (\vh_{\riding} \rightarrow \canride(s)\\ 
	& \qquad \qquad \quad \quad \wedge \isridable(o)))]
\end{align*}
where $y$ is the given training label and $\dvrd$ is the training dataset.  This rule reduces the learning
burden of the classifier for \colons{Riding} class by allowing feedback only when $\canride(s)$ is \TRUE{}.  We
introduce a few more rules by adding them in conjunction with the above rules (see supplementary).  These
rules can be obtained from taxonomies (\eg{} ConceptNet) or meta-data of prior datasets (\eg{} VisualGenome
\cite{krishna2017visual}).

\paragraph{Evaluation:} 
We follow \cite{yu2017visual}, reporting Recall@N (R@N), the recall of the top-N prediction scores in 
an image where we take into account all $70$ predictions per object pair. This strategy is different from \cite{lu2016visual}, which only considers the top prediction for each object pair penalizing cases where multiple predicates apply equally well but were omitted by the annotators. 

%An image with $5$ object pairs will generate $350$
%predictions, $5$ of which are correct.  If $4$ of those $5$ correct predictions appear in the top $100$ scores, then
%R@100 for that image would be $80\%$.

\begin{table*}[htbp!]
	\centering
		\begin{tabular}{cc|cccc} 
		\hline 
		\multicolumn{2}{c|}{Method} & \multicolumn{2}{c}{R@50} & \multicolumn{2}{c}{R@100} \\
		& & Standard  & Zero-Shot & Standard & Zero-Shot \\ \hline 
		\multicolumn{2}{c|}{} & \multicolumn{3}{c}{$1\%$ Data} \\ \hline \hline
		\multicolumn{2}{l|}{VGG (baseline)} & $60.8\pm6.7$ & $40.7\pm5.8$ & $75.4\pm7.8$ & $59.4\pm8.1$  \\
		\multicolumn{2}{l|}{\ \  + Knowledge} & $\quad 68.5\pm1.8^{**}$ & $\quad49.5\pm1.5^{**}$ & $\quad83.1\pm1.6^{**}$ & $\quad70.1\pm2.4^{**}$ \\
		%\multicolumn{2}{l|}{\ \  + Logical + Ext-Odds} &  $78.5\pm0.1$ & $53.4\pm0.2$ & $88.9\pm0.0$ & $73.9\pm0.5$ \\
		\hline
		\multicolumn{2}{l|}{VGG-SS (baseline)} & $67.9\pm8.5$ & $47.6\pm8.5$ & $80.3\pm7.6$ & $65.6\pm9.2$  \\
		\multicolumn{2}{l|}{\ \  + Knowledge} & $\;\:74.0\pm0.7^{*}$ & $\;\:54.4\pm1.4^{*}$ & $\;\:85.9\pm0.5^{*}$ & $\;\:73.4\pm1.2^{*}$ \\
		%\multicolumn{2}{l|}{\ \  + Logical + Ext-Odds} & $79.7\pm0.2$ & $56.4\pm0.4$ & $89.6\pm0.4$ & $75.6\pm0.7$ \\ 
		\hline \hline
	
		\multicolumn{2}{c|}{} & \multicolumn{3}{c}{$5\%$ Data} \\ \hline \hline
		\multicolumn{2}{l|}{VGG (baseline)} & $70.3\pm0.5$ & $48.4\pm1.0$ & $83.5\pm0.4$ & $68.3\pm0.9$   \\
		\multicolumn{2}{l|}{\ \  + Knowledge} & $\quad73.8\pm0.5^{**}$ & $\quad53.4\pm0.9^{**}$ & $\quad86.4\pm0.4^{**}$ & $\quad73.7\pm1.1^{**}$   \\ 
		%\multicolumn{2}{l|}{\ \  + Logical + Ext-Odds} & $80.1\pm0.1$ & $56.0\pm0.5$ & $90.5\pm0.1$ & $77.7\pm0.4$ \\
		\hline 
		\multicolumn{2}{l|}{VGG-SS (baseline)} & $79.6\pm0.4$ & $58.1\pm1.2$ & $89.6\pm0.3$ & $77.1\pm1.1$  \\ 
		\multicolumn{2}{l|}{\ \  + Knowledge} & $79.9\pm0.4$ & $\quad59.6\pm0.9^{**}$ & $89.7\pm0.3$ & $\quad78.5\pm0.8^{**}$   \\ 
		%\multicolumn{2}{l|}{\ \  + Logical + Ext-Odds} & $82.7\pm0.2$ & $61.5\pm0.4$ & $91.8\pm0.3$ & $80.6\pm0.8$ \\ 
	
		\hline
		\end{tabular}
		\caption{Performance on the predicate detection task
		from the Visual Relationship Dataset \cite{lu2016visual} with and without commonsense knowledge. 
		We conduct the experiments in data scarce condition using $1\%$ and $5\%$ training data and
		report Recall@N averaged (with standard deviation) across $10$ random subsets. \colons{VGG} refers to a network using VGG-16 based visual features \cite{liang2018visual} and \colons{VGG-SS} combines semantic and spatial features
		with the visual features. We report the statistical significance between \colons{baseline} and corresponding knowledge augmented model (\colons{+ Knowledge}) ($\text{p-value}<0.01$ as $**$ and $\text{p-value}<0.05$ as $*$).}
			\label{tab:vrd}
	\end{table*}

\paragraph{Results:} \autoref{tab:vrd} shows the results on the VRD dataset
when training with knowledge (\textit{+ Knowledge}) and without knowledge (\textit{baseline}) for the two variants and
for both the standard and zero-shot settings.  We observe consistent improvements across all cases with augmentation of
knowledge.  The improvements are higher for the $1\%$ data ($+7.7\%$ for R@100 for Standard) than the $5\%$ data
($+2.9\%$ for R@100 for Standard) showing that knowledge has more benefits in lower data regimes. We made similar
observation for the MNIST toy example.  The improvements are generally higher for the zero-shot setting ($+10.7\%$ for
R@100 in the $1\%$ case) since this setting is inherently data starved and prior semantic knowledge helps to regularize
the model in such conditions.  We also note that the improvements are comparatively smaller for the VGG-SS network since
semantic and spatial information are being explicitly injected as features into the model. Although there are some
overlaps between the semantic features and provided declarative knowledge, they are fundamentally different, and could
complement each other as observed above ($59.4\%$ of VGG versus $78.5\%$ of VGG-SS + Knowledge).  Our results show that
DASL obtains better generalization in data scarce settings by augmenting the NN with commonsense rules.

\section{Conclusion}
\label{sec:conclusions}

In this paper, we introduced \textbf{Deep Adaptive Semantic Logic (DASL)} to unify machine reasoning and machine learning. DASL is fully general,
encompassing all of first order logic and arbitrary deep learning architectures.  DASL improves deep learning by
supplementing training data with declarative knowledge expressed in first order logic.  The vocabulary of the domain
is realized as a collection of neural networks.  DASL composes these networks into a single DNN and applies
backpropagation to satisfy both data and knowledge.  We provided a formal grounding which demonstrates the correctness
and full generality of DASL for the representation of declarative knowledge in first order logic, including correctness
of mini-batch sampling for arbitrary domains.  This gives us to freedom to apply DASL in new domains without requiring
new correctness analysis.

We demonstrated a $1000$-fold reduction in data requirements on the MNIST digit classification task by using declarative
knowledge in the form of arithmetic relation satisfied by unlabeled image triplets.  The knowledge {\em restricted} the
behavior of the model, preventing erroneous generalization from the small number of labeled data points.  We then
demonstrated the application of commonsense knowledge to visual relationship detection, improving recall from 59.4 to
70.1.  Here, knowledge was used to {\em free} the model from the burden of learning cases covered by the knowledge,
allowing the model to do a better job of learning the remaining cases.

First order logic provides a uniform framework in which we plan to support transfer learning and zero-shot learning by
training DASL models on theories where data is abundant and then creating new theories on the same vocabulary that
address problems where data is sparse.  We also plan to demonstrate the converse capability, training distinct models of
a single theory, allowing us to sample models as a technique for capturing true probabilities, similar to Markov Logic
Networks \cite{richardson2006markov}.  Finally, we are exploring ways to allow DASL to learn rules from data while
retaining explainability and integrating smoothly with user defined logic.

\section{Acknowledgements}

This material is based upon work supported by the Defense Advanced Research Projects Agency (DARPA) under Contract No. HR001118C0023. Any opinions, findings and conclusions or recommendations expressed in this material are those of the author(s) and do not necessarily reflect the views of DARPA. The authors would like to acknowledge Karen Myers, Bill Mark, Rodrigo de Salva Braz, and Yi Yao for helpful discussions.

{\small
\bibliographystyle{icml2020}
\bibliography{egbib}
}

\newpage
\onecolumn

\section{Supplementary Material}

\subsection{Curriculum learning for MNIST Toy Example}

In \textbf{section 4.1} we trained a NN for digit classification on the MNIST dataset in a data scarce setting. We used a few labeled samples and a large number of unlabeled triplets satisfying some rules (modular arithmetic in our experiments).
We used a curriculum based learning strategy to prevent the model from collapsing to a degenerate solution, especially
for cases with extremely small number of labeled samples (\eg{} $2$ samples per class).
In such cases the model tends to get trapped in a local minimum where the axiom corresponding to the unlabeled triplets can be satisfied by a solution with all digits being classified as $0$ since $0 + 0 = 0$.
Within the curriculum, we begin the training with all the labeled examples and a small working set of the unlabeled triplets. We progressively expand the working set during training as the model becomes more confident on the unlabeled examples.
The confidence score $p_c^t$ is computed using a low-pass filter:
\begin{align*}
&p_c^{t} = (1 - \alpha) * p_c^{t-1} + \alpha * p_{max}
\end{align*}
where $*$ is scalar multiplication, $t$ is the iteration index, $p_c^{0}=0$, $\alpha=0.1$, and $p_{max}$ is the average probability of the highest scoring class on the first digit of the triplet. When $p_c^t>0.9$, we increase the working set of unlabeled triplets by a factor of $2$ until it reaches the maximum number of unlabeled triplets. When $p_c^t>0.9$, we reset $p_c^t$ to let the model fit well to the new working set before reaching the condition again. This curriculum ensures that the model is able to find a decent initialization using the labeled examples and then progressively improve using the unlabeled samples. The initial set of unlabeled triplets contained $10$ samples per class and the maximum number of triplets is bounded by the class with minimum number of samples. During the final curriculum step we remove all labeled data, allowing the model to train solely on the rules. This allows the model to trade off errors on the labeled data for better overall performance.

\subsection{Normalized Relative Spatial Features for Visual Relationship Detection}

We provide the implementation details for the spatial features used in the
visual relationship detection experiments in \textbf{section 4.2}.
These features capture the relative spatial configuration of the subject and the object bounding boxes and were used to
augment visual and semantic features for predicting the visual relationship (VGG-SS).
We denote the
coordinates of the object and subject bounding boxes as $(x_s, y_s, w_s, h_s)$ and $(x_o, y_o, w_o, h_o)$ respectively, where $(x, y)$ are the coordinates of the (box) center with width $w$ and height $h$.
The relative normalized features is an eight dimensional feature and computed as
$\left[\frac{x_s-x_o}{w_o}, \frac{y_s-y_o}{h_o}, \frac{x_o-x_s}{w_s}, \frac{y_o-y_s}{h_s}, \log(\frac{w_s}{w_o}), \log(\frac{h_s}{h_o}), \log(\frac{w_o}{w_s}), \log(\frac{h_o}{h_s}) \right]$. These features were also used in the baseline model \cite{liang2018visual}.

\subsection{Commonsense rules for Visual Relationship Detection}
In addition to the rule described in \textbf{section 4.2}, we used additional rules for incorporating commonsense knowledge in predicting visual relationships using DASL. These rules follow the same format as the rule for the \colons{Riding} predicate described earlier and are outlined below:

\begin{enumerate}
    \item \colons{Wear} predicate should only apply when the subject is a \textit{living entity} and the object is \textit{wearable}.
    \item \colons{Sleep-On} predicate should only apply when the subject is a \textit{living entity} and the object is \textit{sleepable}.
    \item \colons{Eat} predicate should only apply when the object is \textit{eatable}.
    \item Predicates- \colons{Above}, \colons{Over}, \colons{Ride}, \colons{On-The-Top-Of}, \colons{Drive-on}, \colons{Park-On}, \colons{Stand-On}, \colons{Sit-On}, \colons{Rest-On} should apply only when the subject is spatially \textit{above} the object. We defined \textit{above} as a function that is \TRUE{} when $y_s \geq y_o$.
    \item Predicates- \colons{Under}, \colons{Beneath}, \colons{Below}, \colons{Sit-Under} should apply only when the subject is spatially \textit{below} the object. We defined \textit{below} as a function that is \TRUE{} when $y_s \leq y_o$.
    \item Predicates- \colons{On-The-Right-Of} should apply only when the subject is spatially \textit{right of} the object. We defined \textit{right of} as a function that is \TRUE{} when $x_s \geq x_o$.
    \item Predicates- \colons{On-The-Left-Of} should apply only when the subject is spatially \textit{left of} the object. We defined \textit{left of} as a function that is \TRUE{} when $x_s \leq x_o$.
\end{enumerate}

These rules cover facts related to both semantic and spatial commonsense knowledge.
We incorporated these rules by adding them in conjunction with original theory presented in \textbf{section 4.2}.

\begin{align*}
	&(\forall (\imag, s, o, y):\dvrd) [\pi_y(\vrd(\imag, s, o) \\
	& \quad \wedge (\vh_{\riding} \rightarrow \canride(s)\\
    & \qquad \qquad \quad \quad \wedge \isridable(o))\\
    & \quad \wedge (\vh_{\wear} \rightarrow \isliving(s)\\
    & \qquad \qquad \quad \quad \wedge \iswearable(o))
    \ldots)]
\end{align*}

where $\vh_{Cls} \in \R^{70}$ is a one-hot vector of truth values,
which is \TRUE{} at the index of the predicate class \colons{Cls} and \FALSE{} elsewhere. $\isliving$
is a constant vector
of truth values for all objects which is \TRUE{} at indices of objects which are living entities and \FALSE{}
elsewhere. Similarly, $\iswearable$ is a constant vector, which is \TRUE{} at
exactly the indices of objects which are wearable. We refer readers to \textbf{section 4.2} for detailed explanation about the application of these rules.

\end{document}